\documentclass[preprint,12pt,3p,times]{elsarticle}

\usepackage{amssymb}
\usepackage{amsmath}

\usepackage{xcolor}

\usepackage{algorithm}
\usepackage[noend]{algpseudocode}
\usepackage{array}
\usepackage{textcomp}
\usepackage{stfloats}
\usepackage{url}
\usepackage{verbatim}
\usepackage{graphicx}
\usepackage{booktabs}
\usepackage{multirow}
\usepackage{setspace}
\usepackage{lipsum}  
\usepackage{textcomp}
\usepackage[caption=false,font=normalsize,labelfont=sf,textfont=sf]{subfig}

\definecolor{olive}{rgb}{0.5, 0.5, 0.0}
\definecolor{maroon}{rgb}{0.69, 0.19, 0.38}
\definecolor{celestialblue}{rgb}{0.29, 0.59, 0.82}
\definecolor{darkgreen}{rgb}{0.0, 0.6, 0.0}
\definecolor{grey}{rgb}{0.5,0.5,0.5}
\definecolor{darkblue}{rgb}{0.19, 0.19, 0.62}
\definecolor{silver}{rgb}{0.7,0.7,0.7}
\definecolor{darkcyan}{rgb}{0.0, 0.55, 0.55}

\newcommand{\myeq}[1]{\hfill{\refstepcounter{equation}(\theequation)\label{#1}}}

\begin{document}

\begin{frontmatter}



\title{Time Frequency Analysis of EMG Signal for Gesture Recognition using Fine-grained Features}


\author[first]{Parshuram N. Aarotale}
\affiliation[first]{organization={Wichita State University},
            addressline={Dept. of Biomedical Engineering}, 
            city={Wichita},
            postcode={67220}, 
            state={Kansas},
            country={USA}}

\author[second]{Ajita Rattani}
\affiliation[second]{organization={University of North Texas},
            addressline={Dept. of Computer Science and Engineering }, 
            city={Denton},
            postcode={}, 
            state={Texas},
            country={USA}}

\begin{abstract}

Electromyography (EMG)–based hand‑gesture recognition converts forearm muscle activity into control commands for prosthetics, rehabilitation, and human–computer interaction. This paper proposes a novel approach to EMG-based hand gesture recognition that uses \emph{fine‑grained} classification and presents \textbf{XMANet}, which unifies low‑level local and high‑level semantic cues through cross‑layer mutual‑attention among shallow‑to‑deep CNN experts. Using stacked spectrograms and scalograms derived from the Short‑Time Fourier Transform (STFT) and Wavelet Transform (WT), we benchmark XMANet against ResNet50, DenseNet‑121, MobileNetV3, and EfficientNetB0.
Experimental results on the Grabmyo dataset indicate that, using STFT, the proposed XMANet model outperforms the baseline ResNet50, EfficientNetB0, MobileNetV3, and DenseNet121 models with improvement of approximately $1.72\%$, $4.38\%$, $5.10\%$, and $2.53\%$, respectively. When employing the WT approach, improvements of around $1.57\%$, $1.88\%$, $1.46\%$, and $2.05\%$ are observed over the same baselines. Similarly, on the FORS EMG dataset, the XMANet(ResNet50) model using STFT shows an improvement of about $5.04\%$ over the baseline ResNet50. In comparison, the XMANet(DenseNet121) and XMANet(MobileNetV3) models yield enhancements of approximately $4.11\%$ and $2.81\%$, respectively. Moreover, when using WT, the proposed XMANet achieves gains of around $4.26\%$, $9.36\%$, $5.72\%$, and $6.09\%$ over the baseline ResNet50, DenseNet121, MobileNetV3, and EfficientNetB0 models, respectively. These results confirm that XMANet consistently improves performance across various architectures and signal processing techniques, demonstrating the strong potential of fine-grained features for accurate and robust EMG classification.
\end{abstract}




\end{frontmatter}



\section{Introduction}
\label{introduction}

Surface electromyogram (sEMG) signals are integral to the advancement of human-machine interface (HMI) systems. Their versatile applications span myoelectric prosthesis control~\cite{atzori2016deep}, rehabilitative feedback systems~\cite{kumar2019prosthetic}, disease prediction frameworks~\cite{muzammal2020multi}, and neurorobotics~\cite{gulati2021toward}. In particular, sEMG-based gesture recognition has emerged as a critical component for assistive technologies in individuals with limb amputations. By enabling the precise decoding of muscle activation patterns, these systems facilitate the translation of user intent into accurate prosthetic hand movements, thereby enhancing both the functional performance of the device and the overall user experience.

Recent advancements in hand gesture recognition have leveraged both conventional and deep learning techniques to enhance accuracy and control. Classical machine learning methods such as k-Nearest Neighbors (KNN), Linear Discriminant Analysis (LDA), and Support Vector Machines (SVM) have been effectively applied to classify hand gestures~\cite{khushaba2012toward,tuncer2022novel,khushaba2012electromyogram,essa2023features,pancholi2018portable}. Furthermore, hybrid approaches that combine fuzzy cognitive maps (FCM) and Bayesian belief networks with extreme learning machines (ELM) have shown considerable promise in gesture classification~\cite{prabhakar2023efficient}. Additionally, feature extraction strategies such as variational mode decomposition (VMD) integrated with a multiclass SVM framework and the application of fractional Fourier transform (FrFT) features with KNN have demonstrated competitive performance in discerning complex hand gestures~\cite{prabhavathy2024hand,taghizadeh2021finger}.

Deep learning approaches have achieved remarkable success in hand gesture recognition. Real-time systems employing deep architectures have reached accuracies of approximately $94\%$, often outperforming traditional models~\cite{lee2021electromyogram}. In particular, Convolutional Neural Networks (CNNs) have been extensively adopted~\cite{asif2020performance,chen2020hand,wang2023hand,zhang2022emg,wang2023emg}, while Long Short-Term Memory (LSTM) networks have demonstrated competitive performance~\cite{ovur2021novel,barron2020recurrent,topalovic2019emg}.

However, cross-study comparisons are often hampered by variations in experimental conditions. Differences in sensor types, sampling frequencies, and placements (e.g., elbow versus forearm) can significantly impact performance metrics, thereby limiting the generalizability of the results~\cite{pancholi2018portable,zhang2022emg,wang2023emg,chamberland2023novel,ghislieri2021long}. Hand gesture recognition using sEMG signals is inherently challenging due to high inter-class similarity and intra-class variability. Conventional CNNs, when applied to representations such as the Short-Time Fourier Transform (STFT) or wavelet transforms, may overlook subtle temporal-frequency nuances essential for fine-grained classification, as they often discard low-level details~\cite{liu2023learn}. A number of studies have proposed fine-grained feature analysis for enhancing the performance of classification tasks in various domains~\cite{maji2013fine,van2017devil,zhao2017survey,yin2024fine,ge2018chest}.
We aim to leverage fine-grained analysis, integrating both high- and low-level features, toward accurate EMG-based hand gesture recognition, XMANet.

To address these challenges, we propose \textbf{XMANet}, a Cross-layer Mutual Attention Learning Network for fine-grained feature classification. In this framework, each convolutional layer acts as a specialized expert, autonomously identifying attention regions that emphasize discriminative features at distinct scales. These attention regions are then exchanged across layers via a mutual learning mechanism, resulting in a comprehensive feature representation that also serves as an effective form of data augmentation. This approach enhances model generalization and robustness, reducing misclassifications among similar gestures.

In summary, proposing XMANet for sEMG-based hand gesture recognition systems holds significant promise for enhancing classification accuracy and interpretability, effectively addressing key challenges in feature extraction and the inherent variability of sEMG signals. Thus, to advance the state-of-the-art in electromyography (EMG)-based gesture recognition, this paper introduces a novel framework that integrates the Fourier Transform and Time-frequency feature extraction with an innovative fine-grained version of deep learning architectures. Our approach, sEMG-XMANet, is designed to address the inherent complexities of EMG signals by leveraging advanced signal processing techniques and a multi-expert deep learning strategy. The primary \textbf{contributions} of our work are summarized as follows:


\vspace{0.25 cm} \noindent\textbf{Our Contribution:}
\begin{itemize}

\item \textbf{Advanced Feature Extraction and Signal Representation:} Innovative techniques for processing raw EMG signals are presented. These include the segmentation and transformation of the signals into sEMG images, along with the extraction of Short-Time Fourier Transform (STFT) stacked spectrograms and time-frequency stacked scalograms, which provide enriched representations of signal dynamics.

\item \textbf{Development of Novel Fine-grained Deep Learning Model (XMANet):} A deep learning model based on cross-layer mutual attention learning is developed. In this architecture, predictions are aggregated from shallow layers (capturing fine, low-level details) to deep layers (capturing high-level semantic features), with each layer treated as an independent expert to improve hand-based gesture recognition performance.

\item \textbf{ Experimental Validation on Latest Datasets:} Comprehensive experiments are conducted on two publicly available latest datasets, namely Grabmyo~\cite{pradhan2022multi} and FORS-EMG~\cite{rumman2024fors}, to validate the effectiveness of the proposed method.

\end{itemize}

This paper is organized as follows. In section $2$, we discuss the related work on EMG-based gesture recognition using machine learning and deep learning models. In section $3$, we discuss feature extraction techniques  and  In section $4$ discuss proposed deep-learning architectures used in this study. In Section $5$, the datasets, data pre-processing and segmentation, and evaluation metrics are discussed. The results are analyzed in section $6$. Conclusions and future directions are discussed in Section $7$.

\section{Related Work}

Recent advancements in surface electromyography (sEMG) for hand gesture recognition have seen the development of various techniques employing both traditional machine learning methods and advanced deep learning models. This section reviews the contributions of several studies in this domain.

\subsection{Traditional Machine Learning Approaches}

Tuncer et al.~\cite{tuncer2022novel} applied a multilevel feature extraction approach combined with fine-tuning of k-Nearest Neighbor (kNN) and Support Vector Machine (SVM) classifiers to distinguish fifteen individual as well as combined finger movements in a publicly available dataset~\cite{khushaba2012electromyogram}. In a similar vein, Provakar et al.~\cite{prabhakar2023efficient} introduced four hybrid models for finger movement classification. Their models included one based on graph entropy, another that utilized fuzzy cognitive maps (FCM) alongside empirical wavelet transformation (EWT), a third that merged fuzzy clustering with a least-squares support vector machine (LS-SVM) classifier, and a fourth that incorporated Bayesian belief networks (BBN) with extreme learning machines (ELM).

In another study, Essa et al.~\cite{essa2023features} evaluated five distinct sets of features using kNN, Linear Discriminant Analysis (LDA), and SVM classifiers to categorize $17$ gestures, with the LDA classifier exhibiting particularly high accuracy. Furthermore, Prabhavathy et al.~\cite{prabhavathy2024hand} proposed an sEMG-based gesture classification framework that employs the Variational Mode Decomposition (VMD) technique to improve hand movement classification accuracy. Their framework used a multi-class SVM with a one-vs-one (OVO) strategy, achieving an average accuracy of $99.98\%$.

Similarly, Heydarzadeh et al.~\cite{heydarzadeh2017emg} reached an accuracy of $89\%$ when classifying ten types of movements using an SVM classifier, with reflection coefficients serving as the feature vector. Their approach also involved estimating the Power Spectral Density (PSD) of sEMG signals from two-channel data with various window widths. Additionally, Subasi et al.~\cite{subasi2022surface} utilized the tunable Q-factor wavelet transform (TQWT) together with bagging and boosting ensemble classifiers to identify six different hand movements.
Finally, Lee et al.~\cite{lee2021electromyogram} developed a real-time gesture recognition system for hand and finger movements based on $18$ time-domain features. Their study revealed that Artificial Neural Networks (ANN) outperformed other algorithms, such as SVM, Random Forest (RF), and Logistic Regression (LR), achieving an accuracy rate of $94\%$.

\subsection{Deep Learning Approaches}

Convolutional Neural Networks (CNNs) have been extensively applied to address hand gesture recognition challenges, as demonstrated by several studies in the literature~\cite{wang2023hand,zhang2022emg,wang2023emg}. For example, Atzori et al.~\cite{atzori2016deep} employed a straightforward CNN architecture to classify nearly $50$ gestures from one of the NinaPro databases, noting that the first layer's configuration, weight initialization, data augmentation methods, and learning rate significantly affected the results. Similarly, Yang et al.~\cite{yang2019emg} investigated the use of raw EMG signals as CNN input by comparing time- and frequency-domain representations on two public datasets, with especially promising outcomes on the CapgMyo-DBa dataset. Asif et al.~\cite{asif2020performance} carried out a detailed analysis of the influence of various hyperparameters, such as learning rate and number of epochs, confirming that these selections are critical to convolutional neural network performance.

Chamberland et al.~\cite{chamberland2023novel} introduced EMaGer, a scalable $64$-channel HD-EMG sensor designed to accommodate different forearm sizes, and leveraged a CNN-based model for gesture classification. 
In addition to CNNs, LSTM networks have been used in several studies. Ghislieri et al.~\cite{ghislieri2021long} demonstrated that LSTM networks can effectively detect muscle activity in EMG recordings without the need for background-noise computations. Moreover, Cai et al.~\cite{cai2021hybrid} extracted both spatial and temporal features by combining CNNs with LSTM networks. Li et al.~\cite{li2022emg} proposed a CNN-LSTM architecture to classify five dynamic hand gestures across different limb positions, revealing that the model's accuracy was affected by limb position, inter-person variability, and the choice of dynamic gestures.

Furthermore, Zhao et al.~\cite{zhang2022emg} incorporated fuzzy logic with LSTM networks to account for temporal correlations across EMG signal windows, successfully classifying up to four hand motions. In contrast, Antonius et al.~\cite{antonius2021electromyography} achieved promising results by merging CNNs with an LSTM-like recurrent neural network, albeit for a smaller set of basic gestures. Karman et al.~\cite{karnam2022emghandnet} employed a bidirectional LSTM together with CNNs to capture both forward and backward inter-channel temporal information. Barona López et al.~\cite{lopez2024cnn} demonstrated that a CNN-LSTM model outperforms a CNN-only approach in EMG-based hand gesture recognition, achieving a post-processed recognition accuracy of $90.55\pm9.45$, compared to $87.26\pm11.14$ for the CNN-only model.

\section{Feature Extraction Techniques: Fourier and Wavelet Transform-Based Spectrogram}

\begin{figure*}[ht]
    \centering
    \includegraphics[width=0.75\linewidth]{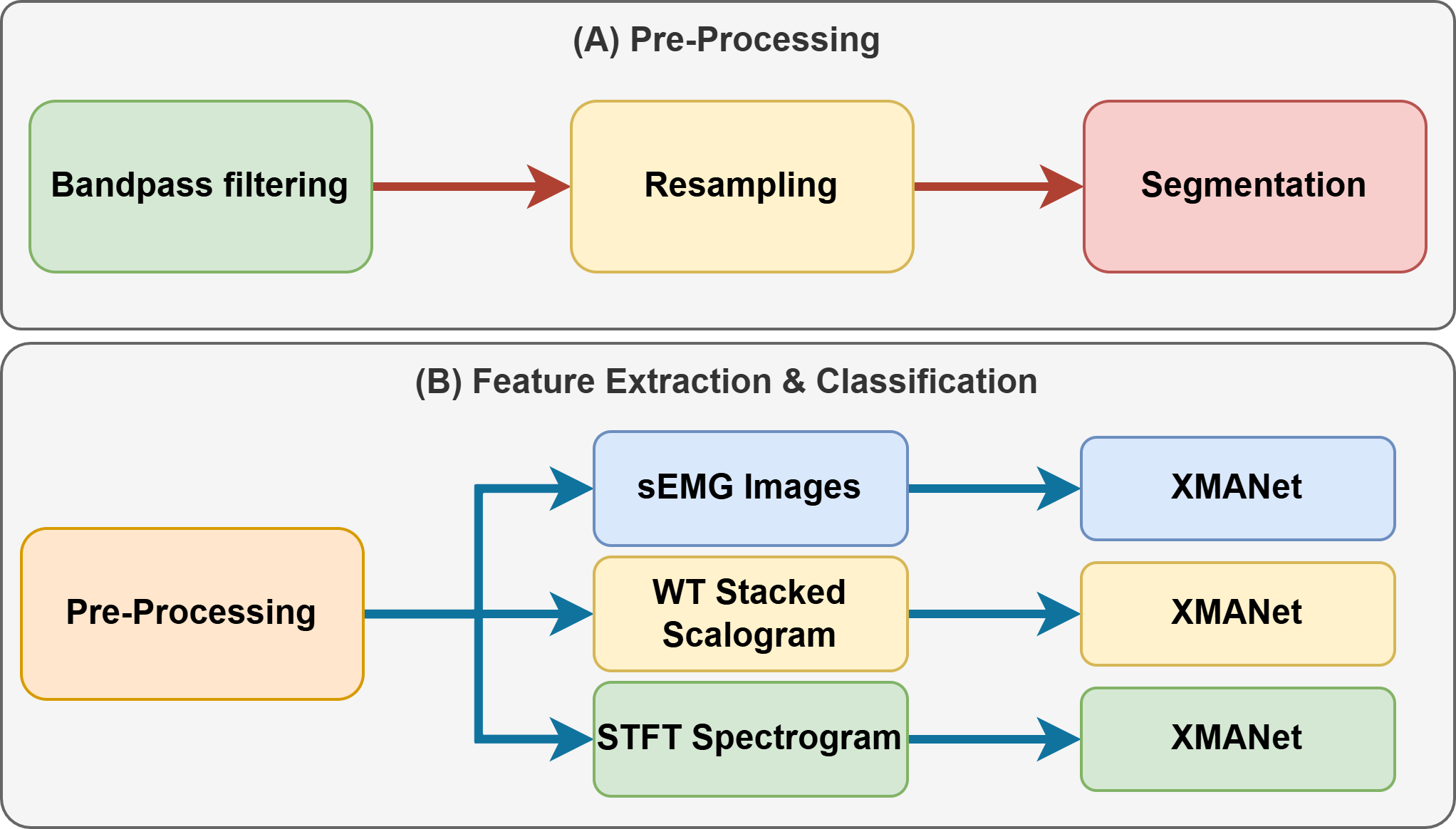}
    \caption{Figure illustrates the outline of proposed Method. A) Pre-Processing, B) Feature Extraction and Classification techniques for Gesture Recognition.}
    \label{fig:blocfeature}
\end{figure*}

This section outlines the feature‑extraction techniques employed, specifically the Short‑Time Fourier Transform (STFT) and Wavelet Transform (WT), which are summarized in Fig.~\ref{fig:blocfeature}.

\subsection{Short Time Fourier Transform Spectrogram}

Each (sEMG) signal segment was transformed into a spectrogram using the Short-Time Fourier Transform (STFT).
The Short-Time Fourier Transform (STFT) of a signal \( x(t) \) is defined as:

\begin{equation}
\text{STFT}\{x(t)\}(m, \omega) = \int_{-\infty}^{\infty} x(t) w(t - m) e^{-j\omega t} dt
\end{equation}

where \( x(t) \) is the input signal, \( w(t) \) is a window function, \( m \) is the time shift, and \( \omega \) is the angular frequency \cite{oppenheim1999discrete}.

The magnitude of the STFT coefficients is used to generate the spectrogram:

\begin{equation}
S(t, f) = \left| \text{STFT}\{x(t)\}(t, f) \right|
\end{equation}

The logarithm of the magnitude is calculated to enhance the visibility of the signal's features:

\begin{equation}
\text{Log-Spectrogram}(t, f) = \log(S(t, f) + 1e-7)
\end{equation}

Normalization is performed to scale the log-spectrogram between $0$ and $1$:

\begin{equation}
S_{\text{norm}}(t, f) = \frac{S(t, f) - S_{\min}}{S_{\max} - S_{\min}}
\end{equation}

where \( S_{\max} \) and \( S_{\min} \) are the maximum and minimum values of the logarithmic spectrogram, respectively \cite{griffin1984signal}.

This normalized log-spectrogram was converted into RGB images using a colormap and resized to $224\times224$ pixels to match the input size of pre-trained convolutional neural networks~(CNNs). The individual channel spectrograms were stacked row-wise to form a single Spectrogram image.

\subsection{Wavelet Transform based stacked scalogram}

Each segmented signal was transformed into a scalogram using a continuous wavelet transform (CWT) with the Mexican hat (Mexh) wavelet.

The wavelet transform is a mathematical technique that decomposes a signal into components at various scales and positions, offering a time-frequency representation of the signal. The continuous wavelet transform (CWT) of a signal \( x(t) \) is defined as

\begin{equation}
\text{CWT}(a, b) = \frac{1}{\sqrt{a}} \int_{-\infty}^{\infty} x(t) \psi^*\left(\frac{t - b}{a}\right) dt
\end{equation}

where \( \psi \) is the mother wavelet, \( a \) is the scale parameter, \( b \) is the translation parameter, and \( \psi^* \) denotes the complex conjugate of \( \psi \)\cite{daubechies1992ten,mallat1999wavelet}. The Mexican hat wavelet used in this study is a second derivative of the Gaussian function, providing good localization in both time and frequency domains. The scales ranged from 1 to 128 to capture the frequency components of the sEMG signals.

The power spectrum of the wavelet coefficients was calculated and normalized to enhance the signal's features. The power spectrum \( P(a, b) \) is given by:

\begin{equation}
P(a, b) = \left| \text{CWT}(a, b) \right|^2
\end{equation}

Normalization was performed to scale the power spectrum between 0 and 1, which is expressed as:

\begin{equation}
P_{\text{norm}}(a, b) = \frac{P(a, b) - P_{\min}}{P_{\max} - P_{\min}}
\end{equation}

where \( P_{\max} \) and \( P_{\min} \) are the maximum and minimum values of the power spectrum, respectively~\cite{addison2017illustrated}.
This normalized power spectrum was then converted into RGB images using a colormap, and resized to $224\times224$ pixels to match the input size of pre-trained convolutional neural networks (CNNs). The individual channel scalogram were stacked row-wise to form a single scalogram image.

\begin{figure*}[ht]
    \centering
    \includegraphics[width=1\linewidth]{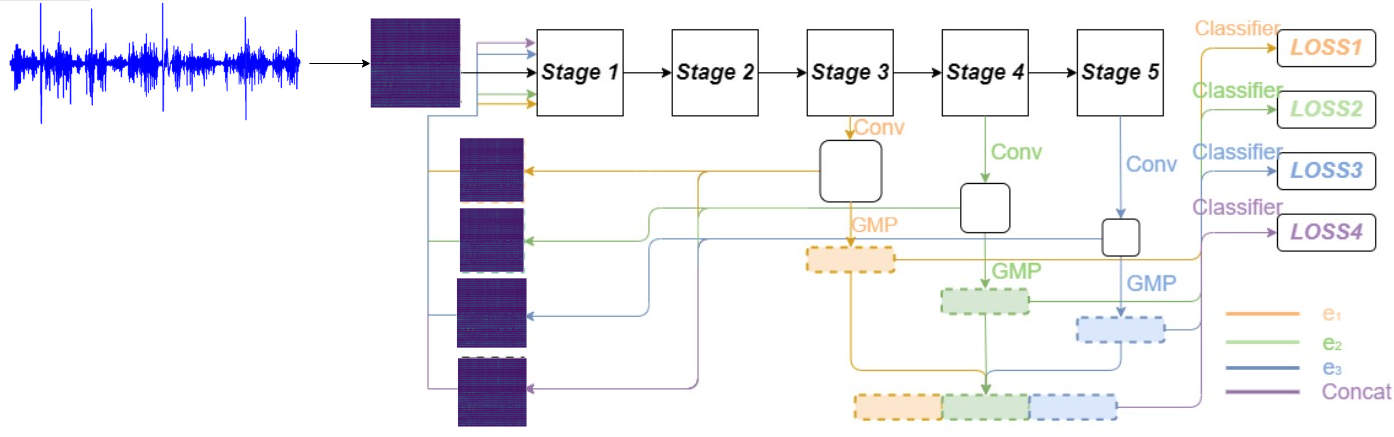}
    \caption{ This figure illustrates XMANet method by introducing three experts $e_{1}$, $e_{2}$, $e_{3}$, on a 5-stage backbone CNN (e.g., ResNet50). The working of each expert and the concatenation of experts are depicted in different colors. Each expert receives feature maps from a specific layer as input and generates a categorical prediction along with an attention region, which is used for data augmentation by other experts. This architecture is trained in multiple steps within each iteration. We start by training the deepest expert (e3), followed by the shallower experts. Finally, in the last step, we train the concatenation of experts to enhance overall performance. 
    }
    \label{fig:my_architecture}
\end{figure*}

\section{Proposed Advanced Fine-grained Features based Methodology (XMANet)}\label{secpm}

In this section, we detail our proposed deep learning framework that exploits multi-level feature representations extracted from lower to upper layers of the CNN model. By incorporating an attention mechanism and a mutual learning strategy, our model refines feature selection and enhances learning efficacy. This approach is developed in line with the seminal contributions in fine-grained classification~\cite{zeiler2014visualizing,jiang2021layercam,liu2023learn,manzoor2024fineface}.

\subsection{Expert Construction: Using Shallow to Deep Layers}


In this part, we describe how a series of \emph{experts} are derived by progressively stacking layers from a chosen backbone CNN. Any modern architecture such as ResNet50, DenseNet121, and related variants may serve as the backbone, denoted by $\beta$. Suppose $\beta$ comprises $M$ convolutional layers indexed as $\{l_{1},l_{2},\dots ,l_{m},\dots ,l_{M}\}$, ordered from the shallowest to the deepest (fully connected layers are excluded).

We define a collection of $N$ experts $\{e_{1},e_{2},\dots ,e_{n},\dots ,e_{N}\}$, where each expert $e_{n}$ aggregates all layers from $l_{1}$ through some $l_{m_{n}}$ with $1\le m_{n}\le M$. Consequently, the experts form a hierarchy that incrementally incorporates deeper layers, and the final expert $e_{N}$ spans the complete stack $l_{1}$ to $l_{M}$.

Let the feature maps output by these experts be $\{x_{1},x_{2},\dots ,x_{n},\dots ,x_{N}\}$, where each $x_{n}\!\in\!\mathbb{R}^{H_{n}\times W_{n}\times C_{n}}$ and $H_{n},W_{n},C_{n}$ denote its spatial height, width, and channel count, respectively. A set of operations $\{F_{1}(\cdot),F_{2}(\cdot),\dots ,F_{n}(\cdot),\dots ,F_{N}(\cdot)\}$ compress these feature maps into fixed‑length, one‑dimensional descriptors $\{v_{1},v_{2},\dots ,v_{n},\dots ,v_{N}\}$, where each $v_{n}\!\in\!\mathbb{R}^{C_{v}}$:

$v_{n}$ = $F_{n}$($x_{n}$) = $f^{GMP}$($x_{n}^{''}$), \myeq{1}\\

$x_{n}^{''}$ = $f^{Elu}$($f^{bn}$($f_{3\times3\times C_{v/2}\times C_{v}}^{conv} (x_{n}^{'})))$, \myeq{2}\\

$x_{n}^{'}$ = $f^{Elu}$($f^{bn}$($f_{1\times1\times C_{n}\times C_{v/2}}^{conv} (x_{n})))$, \myeq{3}\\

Here, $f^{GMP}(\cdot)$ represents global max pooling, $f^{conv}(\cdot)$ a 2‑D convolution parameterized by its kernel size, $f^{bn}(\cdot)$ batch normalization, and $f^{Elu}(\cdot)$ the ELU activation. The intermediate maps $x_{n}^{'}$ and $x_{n}^{''}$ are also utilized to derive the attention region for each expert, as detailed in Subsection~\ref{subsecarp}.

Each expert produces a prediction score $p_{n}=f_{n}^{clf}(v_{n})$ through its dedicated fully connected classifier $f_{n}^{clf}(\cdot)$. In addition to these individual predictions, we construct a holistic descriptor by concatenating all expert vectors:
\[
v_{oval}=f_{concat}(v_{1},v_{2},\dots ,v_{n},\dots ,v_{N}),
\]
followed by a final classifier
\[
p_{oval}=f_{oval}^{clf}(v_{oval}),
\]
which yields the overall prediction score for the framework.

\vspace{-4pt}

\subsection{Attention Region Prediction}\label{subsecarp}

Recall that $x_{n}^{''}$ is the intermediate feature map produced by the $n$‑th expert $e_{n}$.  We consider a $K$‑class classification task, and denote by $k_{n}\!\in\!\{1,2,\dots ,K\}$ the category predicted by $e_{n}$.  The tensor $x_{n}^{''}\!\in\!\mathbb{R}^{H_{n}\times W_{n}\times C_{n}}$ serves as the basis for generating an attention region via a class‑activation map (CAM) that highlights the discriminative image area for class $k_{n}$.

\paragraph{Class‐activation map}
For expert $e_{n}$, the CAM $\Omega_{n}\!\in\!\mathbb{R}^{H_{n}\times W_{n}}$ is computed as
$\Omega_{n}$($\alpha,\beta$) = $\sum\limits_{c=1}^{c_{v}} p_{n}^{c}x_{n}^{''c}(\alpha, \beta)$, 
where $(\alpha,\beta)$ indexes spatial positions, $x_{n}^{''c}$ is the $c$‑th channel of $x_{n}^{''}$, and $p_{n}^{c}$ is the parameter of the classifier $f_{n}^{clf}(\cdot)$ associated with class $k_{n}$.

\paragraph{Upsampling and normalization}
The map $\Omega_{n}$ is bilinearly upsampled to the input image resolution, yielding $\tilde{\Omega}_{n}\!\in\!\mathbb{R}^{H_{in}\times W_{in}}$ (with $H_{in},W_{in}$ the input height and width).  We then apply min–max normalization:\\

$\tilde{\Omega}_{n}^{norm}$($\alpha$, $\beta$) = {$\tilde{\Omega}_{n}$($\alpha$, $\beta$) - min($\tilde{\Omega}_{n}$)}/{max($\tilde{\Omega}_{n}$) - min($\tilde{\Omega}_{n}$)}. \myeq{5}\\

\paragraph{Attention mask and augmentation}
Pixels with positive values in the normalized map $\tilde{\Omega}_{n}^{\text{norm}}$ define a binary mask $\tilde{\Omega}_{n}^{\text{mask}}$.  The corresponding region is cropped from the original image, resized back to $H_{in}\!\times\!W_{in}$, and denoted $A_{n}$.  This $A_{n}$ acts both as the attention region predicted by $e_{n}$ and as an augmented sample for the remaining experts.

\paragraph{Global attention.}
Finally, an overall attention map $A_{oval}$ is obtained by summing the attention information contributed by all experts, thus capturing multi‑level cues across the hierarchy.

\subsection{Multi‑step Mutual Learning}

The entire ensemble is optimized with a progressive \mbox{$N\!+\!2$‑step} schedule, each step minimizing the cross‑entropy loss.  
In the first $N$ stages the experts are updated one at a time from deepest to shallowest so that each model can refine the visual cues lying within its own receptive‑field scope before being influenced by its peers.  
The final two stages bring the experts together: first around the shared attention regions, then around the original image.

\paragraph{Step~1 (deepest expert)}
The iteration begins by training the deepest expert $e_{N}$.  
Because $e_{N}$ subsumes all shallower layers, this pass also yields the set of attention regions
\[
\{A_{1},A_{2},\dots ,A_{n},\dots ,A_{N},A_{oval}\},
\]
which make explicit the “specialized knowledge’’ on which each expert bases its decision.

\paragraph{Steps~2\,–\,$N$ (progress toward shallow)}
For each subsequent expert we move upward in the hierarchy.  
At every step a training sample is drawn at random from a pool that contains the raw image and the attention crops proposed by \emph{other} experts.  
Thus shallower experts benefit from the semantic cues revealed by deeper models, whereas deeper experts can revisit lower‑level structure highlighted by their shallow counterparts.

\paragraph{Step~$N\!+\!1$ (joint training on $A_{oval}$)}
All experts together with their concatenated representation are optimized in a single forward pass using the overall attention region $A_{oval}$.  
This cooperative update encourages the ensemble to integrate the diverse attention information it has accumulated and to focus on still finer‑grained patterns.

\paragraph{Step~$N\!+\!2$ (joint training on raw input)}
Finally, the concatenated descriptor is trained once more on the unaltered image so that the parameters of $f_{oval}^{clf}(\cdot)$ are well aligned with the full‑resolution input, completing the mutual‑learning cycle for that iteration.

\noindent\textbf{Inference Phase:} Figure~\ref{fig:my_architecture} presents the prediction pipeline of the proposed model, which is equipped with $N\!+\!1$ classifiers. Given a test image, the network first outputs $N\!+\!1$ prediction scores. The image’s overall attention region is then passed through the same network, generating another $N\!+\!1$ scores. In total, $2\times(N\!+\!1)$ scores are produced, and their arithmetic mean is taken as the final decision. This two‑pass strategy improves classification accuracy by harnessing two complementary sources of evidence: (a) the ensemble of expert‑specific and global predictions, and (b) features obtained from both the raw input and its corresponding attention map.

\section{Experimental Validation}

\subsection{Datasets}

\begin{figure*} [ht!]
    \centering
    \includegraphics[width=0.92\linewidth]{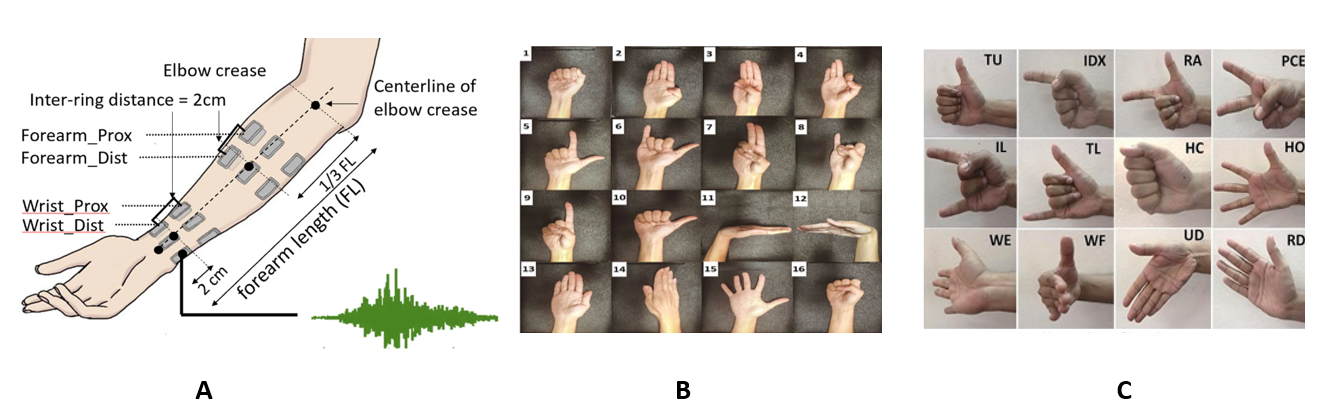}
    \caption{A) Electrode Positions~\cite{pradhan2022multi} and B) Gesture list for Grabmyo dataset~\cite{pradhan2022multi} and  C) FORS-EMG dataset~\cite{rumman2024fors}.}
    \label{figdataset}
\end{figure*}

\noindent \textbf{Grabmyo}~\cite{pradhan2022multi}: This dataset involves $43$ healthy participants ($23$ males and $20$ females) recruited from the University of Waterloo, with data collected on day $1$, day $8$, and day $29$. The participants had an average age of $26.35$ years (±2.89) and an average forearm length of $25.15$ cm (±1.74 cm). Exclusion criteria included muscle pain, skin allergies, or inability to complete all sessions. The study followed ethical guidelines, with participants providing informed consent and the study being approved by the University of Waterloo’s Office of Research Ethics ($31346$).

The experimental setup included a PC and monitor positioned in front of a height-adjustable chair. The data was collected using an EMGUSB2+ amplifier (OT Bioelettronica, Italy) with a gain of $500$ and a sampling rate of $2048$ Hz, along with pre-gelled, skin-adhesive monopolar sEMG electrodes. Participants performed gestures at a normal force level, determined through trial contractions at soft, medium, and hard levels, while following visual instructions on the screen.

This dataset comprises $16$ gestures, including lateral prehension, thumb adduction, various finger oppositions and extensions, wrist flexion and extension, forearm supination and pronation, hand open, and hand close. Gestures were performed in a randomized order with a ten-second rest between each. Each session included seven runs of $17$ gestures (including rest), totaling $119$ contractions per session. This protocol was repeated on day $8$ and day $29$.

\noindent{\textbf{FORS-EMG}~\cite{rumman2024fors}:}
The FORS-EMG dataset used in this work was collected from $19$ healthy individuals who performed $12$ different wrist and finger motions in three different forearm orientations: pronation, neutral (rest), and supination. The participants performed five repetitions of each gesture (see Fig~\ref{figdataset} C) while the electrodes were positioned along the mid-forearm and close to the elbow. For each gesture, eight channels (four on the forearm and four around the elbow) were used to record sEMG signals at $8$ second intervals with a sampling frequency of $985$ Hz. This work processes only the dataset with orientation at rest.

\subsection{Preprocessing and implementation details}

Surface electromyography (sEMG) signals were bandpass filtered with a frequency of $20-450$Hz. Preprocessed signals were segmented into overlapping $0.6$-second windows with a $50\%$ overlap for feature extraction methods such as STFT Stacked Spectrogram and WT stacked Scalogram images. The data sets were divided into $70\%$ for training and $15\%$ each for validation and  testing, and the results were reported on the test set for all experiments.

Data augmentation techniques, such as random resizing, cropping, and horizontal flipping, were applied during training to enhance the model's robustness. Four pre-trained CNN architectures ResNet50, MobileNetV3, and DenseNet121 were fine-tuned for gesture classification by adjusting their final layers to output 17 classes for Grabmyo and 12 classes for Fors EMG dataset respectively. 

Training was conducted using the Adam optimizer and cross-entropy loss function, with early stopping to prevent overfitting. The best-performing model based on validation loss was saved and evaluated on the test set, yielding accuracy metrics and detailed classification reports.

\subsection{Performance Metrics}

The following standard performance metrics ~\cite{fawcett2006introduction} were used for the evaluation of the proposed models in this study. 

\begin{align}
\text{Accuracy} &= \frac{T_{p} + T_{n}}{T_{p} + T_{n} + F_{p} + F_{n}} \\
\text{Precision} &= \frac{T_{p}}{T_{p} + F_{p}} \\
\text{Recall} &= \frac{T_{p}}{T_{p} + F_{n}} \\
F1\text{-score} &= 2 \times \frac{\text{Precision} \times \text{Recall}}{\text{Precision} + \text{Recall}}
\end{align}

where $T_{p}$, $T_{n}$, $F_{p}$, and $F_{n}$ represent true positive, true negative, false positive, and false negative recognition for the given class, respectively.

\section{Results and Discussion}
In this section, we discuss the results obtained by our proposed XMANet in conjunction with different backbones. For instance, XMANet (ResNet50) denotes performance of XMANet using ResNet50 backbone.  The performance of XMANet is compared with existing CNN-based baselines.
\subsection{Experimental Results on Grabmyo}

\begin{table}[ht]
\centering
\caption{Gesture recognition for EMG segmented images using Deep learning CNN-based baseline and the proposed XMANet.}
\label{tab:emgimage}
\resizebox{\columnwidth}{!}{%
\begin{tabular}{l|c|c|c|c}
\hline
\textbf{Models} & \textbf{ACC} & \textbf{P} & \textbf{R} & \textbf{F1} \\ \hline
ResNet50   & 99.18  &  0.99    & 0.99   &   0.99          \\
MobilNetV3   &   99.26 &   0.99   &  0.99  &  0.99         \\
DenseNet121  &       99.44    &   0.99         &   0.99          &     0.99        \\
EfficientNetB0  &       99.12     &   0.99          &   0.99          &     0.99        \\ 
XMANet(ResNet50)  &      \textbf{ 99.60 }     &   1.0          &   1.0        &     1.0        \\
XMANet(DensNet121)  &      \textbf{ 99.61 }    &   1.0          &   1.0          &     1.0       \\
XMANet(MobilnetV3)  &      \textbf{ 99.61 }     &   1.0          &   1.0        &     1.0       \\
XMANet(EfficientNetB0)  &       99.47     &   0.99          &   0.99          &     0.99       \\
\hline
\end{tabular}%
}
\end{table}

The proposed XMANet leverages each pretrained CNN backbone ResNet50, DenseNet121, MobileNetV3, and EfficientNetB0 as independent specialized feature extractors. Table~\ref{tab:emgimage} summarizes the performance metrics for various deep learning models applied to gesture recognition on EMG-segmented images. Notably, the DenseNet121 model achieves an accuracy of $98.44\%$, with precision, recall, and F1 score all at $0.99$,Additionally, other pretrained models demonstrate impressive performance with accuracies of $99.18\%$ (ResNet50), $99.26\%$ (MobileNetV3), and $99.12\%$ (EfficientNetB0). Furthermore, the proposed XMANet variants namely XMANet(ResNet50), XMANet(DenseNet121), XMANet(MobileNetV3), and XMANet(EfficientNetB0) deliver enhanced accuracies of $99.60\%$, $99.61\%$, $99.61\%$, and $99.47\%$, respectively, compared to their baseline counterparts.


The proposed \textbf{XMANet(ResNet50)} improved accuracy by $0.424\%$ relative to the baseline ResNet50, by $0.342\%$ over MobileNetV3, by $0.161\%$ over DenseNet121, and by $0.484\%$ over EfficientNetB0.  Likewise, \textbf{XMANet(DenseNet121)} and \textbf{XMANet(MobileNetV3)} each achieved improvement of roughly $0.433\%$, $0.352\%$, $0.171\%$, and $0.494\%$ compared with the respective baselines of ResNet50, MobileNetV3, DenseNet121, and EfficientNetB0. 

Finally,\textbf{XMANet(EfficientNetB0)} yields improvements of $0.292\%$, $0.21\%$, $0.03\%$, and $0.353\%$ over its ResNet50, MobileNetV3, DenseNet121, and EfficientNetB0 counterparts, respectively. These consistent, though marginal, improvements underscore the efficacy of the proposed modifications in refining model performance.

\begin{table}[ht]
\centering
\caption{Gesture recognition from STFT stacked spectrogram Images using Deep learning baselines and proposed XMANet.}
\label{tab:stftimage}
\resizebox{\columnwidth}{!}{%
\begin{tabular}{l|c|c|c|c}
\hline
\textbf{Models} & \textbf{ACC} & \textbf{P} & \textbf{R} & \textbf{F1} \\ \hline
ResNet50   & 91.46  &  0.91    & 0.91   &   0.91         \\
MobilNetV3   &   85.09 &   0.85   &  0.85  &  0.85         \\
DenseNet121  &       91.17    &   0.91         &   0.91          &     0.91        \\
EfficientNetB0  &       88.04     &   0.88          &   0.88         &     0.88        \\ 
XMANet(ResNet50)  &      \textbf{ 93.03 }     &  0.93          &   0.93        &     0.93       \\
XMANet(DenseNet121)  &      \textbf{ 93.48 }    &   0.93         &   0.93         &     0.93       \\
XMANet(MobilNetV3)  &      \textbf{ 89.43 }     &  0.89          &   0.89        &     0.89       \\
XMANet(EfficientNetB0)  &      \textbf{ 91.89}    &   0.92          &   0.92          &     0.92       \\
\hline

\end{tabular}%
}
\end{table}

Table~\ref{tab:stftimage} presents the performance metrics for various deep learning models applied to gesture recognition using STFT stacked spectrogram images. The experimental results demonstrate that the baseline models achieved accuracies of $91.46\%$ (ResNet50), $88.04\%$ (EfficientNetB0), $85.09\%$ (MobileNetV3), and $91.17\%$ (DenseNet121). With the proposed XMANet approach, these accuracies increased to $93.03\%$ XMANet(ResNet50), $91.89\%$ XMANet(EfficientNetB0), $89.43\%$ XMANet(MobileNetV3), and $93.48\%$ XMANet(DenseNet121), respectively. These enhancements underscore the efficacy of the proposed method in boosting model performance.

A detailed comparative analysis shows that the XMANet(ResNet50) model ($93.03\%$) outperforms the baseline ResNet50, EfficientNetB0, MobileNetV3, and DenseNet121 models with improvements of approximately $1.72\%$, $5.67\%$, $9.33\%$, and $2.04\%$, respectively. In a similar manner, the XMANet(EfficientNetB0) model ($91.89\%$) exhibits enhancements of $0.47\%$ compared to the baseline ResNet50, $4.38\%$ relative to the baseline EfficientNetB0, $7.99\%$ over the baseline MobileNetV3, and $0.79\%$ compared to the baseline DenseNet121. Further comparisons indicate that the XMANet(MobileNetV3) model ($89.43\%$) achieves a $1.58\%$ improvement relative to the baseline EfficientNetB0 and a $5.10\%$ gain over the baseline MobileNetV3. Additionally, the XMANet(DenseNet121) model ($93.48\%$) demonstrates gains of $2.21\%$, $6.18\%$, $9.87\%$, and $2.53\%$ over the baseline ResNet50, EfficientNetB0, MobileNetV3, and DenseNet121 models, respectively. These detailed percentage improvements confirm the robustness and effectiveness of the proposed modifications across the different models.

In \textbf{summary}, XMANet model outperforms the baseline ResNet50, EfficientNetB0, MobileNetV3, and DenseNet121 models with improvements of approximately $1.72\%$, $4.38\%$, $5.10\%$, and $2.53\%$, respectively.

\begin{table}[ht]
\centering
\caption{Gesture recognition from Wavelet transform based stacked scalogram Images using Deep learning and proposed XMANet.}
\label{tab:WTimage}
\resizebox{\columnwidth}{!}{%
\begin{tabular}{l|c|c|c|c}
\hline
\textbf{Models} & \textbf{ACC} & \textbf{P} & \textbf{R} & \textbf{F1} \\ \hline
ResNet50   & 97.71  &  0.98    & 0.98  &   0.98          \\
MobilNetV3   &   97.53 &   0.98   &  0.98  &  0.98         \\
DenseNet121  &      97.29    &   0.98        &   0.98          &     0.98        \\
EfficientNetB0  &     97.87     &   0.98          &   0.98         &     0.98        \\  
XMANet(Resnet50)  &      \textbf{ 99.24}     &     0.99      &    0.99       &       0.99     \\
XMANet(DensNet121)  &      \textbf{ 99.30}    &    0.99      &    0.99       &   0.99       \\
XMANet(MobilnetV3)  &      \textbf{ 99.36 }     &   0.99       &    0.99      &    0.99       \\
XMANet(EfficientNetB0)  &      \textbf{ 99.28}    &   0.99        &    0.99       &   0.99        \\
\hline
\end{tabular}%
}
\end{table}

Table~\ref{tab:WTimage} presents the performance metrics for various deep learning models applied to gesture recognition using wavelet transform-based stacked scalogram images. The experimental results show that the baseline models achieved accuracies of $97.71\%$ (ResNet50), $97.87\%$ (EfficientNetB0), $97.53\%$ (MobileNetV3), and $97.29\%$ (DenseNet121). The findings further demonstrate that the proposed approach significantly enhances performance over these baselines. Specifically, the proposed XMANet(ResNet50), XMANet(MobileNetV3), XMANet(DenseNet121), and XMANet(EfficientNetB0) models achieved accuracies of $99.24\%$, $99.36\%$, $99.30\%$, and $99.28\%$, respectively.

A detailed comparative analysis shows that the XMANet(ResNet50) model  outperforms the baseline ResNet50, EfficientNetB0, MobileNetV3, and DenseNet121 models with improvements of approximately $1.57\%$, $1.40\%$, $1.75\%$, and $2.0\%$, respectively.

In a similar manner, the XMANet(EfficientNetB0) model exhibits enhancements of $1.61\%$,$1.44\%$,
$1.79\%$,and $2.05\%$ compared to the baseline ResNet50,EfficientNetB0, MobileNetV3 
and DenseNet121 respectively. 

Further comparisons indicate that the XMANet(MobileNetV3) model exhibits enhancements of $1.69\%$ compared to the baseline ResNet50, $1.52\%$ relative to the baseline EfficientNetB0, $1.88\%$ over the baseline MobileNetV3, and $2.13\%$ compared to the baseline DenseNet121. 

Additionally, the XMANet(DenseNet121) model demonstrates gains of $1.63\%$, $1.46\%$, $1.82\%$, and $2.07\%$ over the baseline ResNet50, EfficientNetB0, MobileNetV3, and DenseNet121 models, respectively. In \textbf{summary}, the proposed models show improvements of approximately $1.57\%$, $1.88\%$, $1.46\%$, and $2.05\%$ over the baseline ResNet50, MobileNetV3, DenseNet121, and EfficientNetB0 models, respectively.

The proposed XMANet enhances EMG-based gesture recognition from time-frequency (TF) images by introducing a cross-layer mutual attention mechanism. Unlike traditional CNNs that emphasize deeper semantic features, XMANet treats each CNN layer from shallow to deep as an expert capturing unique information at different abstraction levels. Shallow layers extract fine-grained frequency and temporal details, while deeper layers capture high-level semantics. Through mutual attention learning, these layers exchange attention-guided insights, enabling targeted augmentation and reducing irrelevant information. This collaboration improves feature discrimination, reduces intra-class variability and inter-class similarity, and increases robustness to real-world challenges like electrode displacement and muscle fatigue, significantly boosting recognition accuracy.

\subsection{Experimental Results for FORS-EMG}

\begin{table}[ht]
\centering
\caption{Gesture recognition for EMG-segmented Image using Deep CNN-based baseline and the proposed XMANet.}
\label{tab:emgimagefors}
\resizebox{\columnwidth}{!}{%
\begin{tabular}{l|c|c|c|c}
\hline
\textbf{Models} & \textbf{ACC} & \textbf{P} & \textbf{R} & \textbf{F1} \\ \hline
ResNet50   & 92.11  &  0.92    & 0.92   &   0.92          \\
MobilNetV3   &   91.1 &   0.91   &  0.91  &  0.91        \\
DenseNet121  &       94.01    &   0.94         &   0.94          &     0.94        \\
EfficientNetB0  &       92.05     &   0.92         &   0.92          &     0.92       \\ 
XMANet(ResNet50)  &      \textbf{ 93.3 }     &   0.93          &   0.93        &     0.93       \\
XMANet(DensNet121)  &      \textbf{ 94.47 }    &   0.94        &   0.94          &    0.94       \\
XMANet(Mobilnetv3)  &      \textbf{ 93.13 }     &   0.93          &   0.93        &     0.93       \\
XMANet(EfficientNetB0)  &      \textbf{ 93.22 }    &   0.93        &   0.93          &    0.93       \\
\hline
\end{tabular}%
}
\end{table}

Table ~\ref{tab:emgimagefors} presents the performance metrics for various deep learning models applied to gesture recognition for EMG segmented images tested on FORS EMG Dataset. 
In our experimental evaluation, the baseline models ResNet50, Mobilnetv3, Densenet121, and EfficientNetB0 achieved accuracies of $92.11\%$, $91.10\%$, $94.01\%$, and $92.05\%$, respectively. With the proposed XMANet, model ResNet50 improved to $93.30\%$ and model DenseNet121 to $94.47\%$, indicating a improvement in  performance for Proposed XMANet models.

A detailed comparative analysis shows that the XMANet(ResNet50) model outperforms the baseline ResNet50, EfficientNetB0,  and MobileNetV3 models with improvements of approximately $1.29\%$, $1.36\%$, and $2.41\%$, respectively. In a similar manner, the XMANet(EfficientNetB0) model exhibits enhancements of $1.20\%$ compared to the baseline ResNet50, $1.27\%$ relative to the baseline EfficientNetB0, $2.33\%$ over the baseline MobileNetV3. Further comparisons indicate that the XMANet(MobileNetV3) model exhibits enhancements of $1.11\%$ compared to the baseline ResNet50, $1.17\%$ relative to the baseline EfficientNetB0, $2.23\%$ over the baseline MobileNetV3. Additionally, the XMANet(DenseNet121) model demonstrates gains of $2.56\%$, $2.63\%$, $3.70\%$, and $0.49\%$ over the baseline ResNet50, EfficientNetB0, MobileNetV3, and DenseNet121 models, respectively.

In \textbf{summary}, the proposed XMANet(ResNet50) model shows improvement of approximately $1.29\%$ over baseline ResNet50. Similarly, the proposed  XMANet(DenseNet121) model exhibits an enhancement of about a $0.49\%$ over baseline DenseNet121 . Proposed  XMANET(MobileNetV3) exhibits an enhancement of about a $2.23\%$ over baseline MobileNetV3. 

The proposed  XMANET(EfficientNetB0)  exhibits an enhancement of about a $1.27\%$ over baseline EfficientNetB0 .These improvements, though modest, underscore the effectiveness of the proposed XMANet in enhancing model performance.

\begin{table}[ht]
\centering
\caption{Gesture recognition from STFT stacked spectrogram Images using Deep learning CNN-based baseline and the proposed XMANet.}
\label{tab:stftimagefors}
\resizebox{\columnwidth}{!}{%
\begin{tabular}{l|c|c|c|c}
\hline
\textbf{Models} & \textbf{ACC} & \textbf{P} & \textbf{R} & \textbf{F1} \\ \hline
ResNet50   & 52.95  &  0.54    & 0.54   &   0.54         \\
MobilNetV3   &   41.28 &   0.41   &  0.41  &  0.41         \\
DenseNet121  &       54.25    &   0.54        &   0.54         &     0.54       \\
EfficientNetB0  &       48.18     &   0.48          &   0.48         &     0.48        \\ 
XMANet(ResNet50)  &      \textbf{ 55.62 }     &  0.56          &   0.56        &     0.56       \\
XMANet(DensNet121)  &      \textbf{ 56.48}    &   0.56        &   0.57         &     0.57      \\
XMANet(Mobilnetv3)  &      \textbf{42.44 }     &  0.43          &   0.42       &     0.42       \\
XMANet(EfficientNetB0)  &      \textbf{48.08}    &   0.48        &   0.48         &     0.47      \\
\hline
\end{tabular}%
}
\end{table}

Table~\ref{tab:stftimagefors} presents the performance metrics for various deep learning models applied to gesture recognition using STFT-stacked Spectrogram images tested on the FORS EMG Dataset. 

In our experimental evaluation, the baseline models  ResNet50, MobileNetv3, DenseNet121, and EfficientNetB0  achieved accuracies of $52.95\%$, $41.28\%$, $54.25\%$, and $48.18\%$, respectively. 

The proposed XMANET achieve enhance accuracy of $55.62\%$ ,$42.44\%$, and $56.48\%$ as compared to baselines ResNet50, MobileNetV3, DenseNet121 and EfficientNetB0 respectively. A detailed comparative analysis shows that the XMANet(ResNet50) model  outperforms the baseline ResNet50, EfficientNetB0, MobileNetV3, and DenseNet121 models with improvements of approximately $5.04\%$, $15.44\%$, $34.73\%$, and $2.53\%$, respectively. 

In a similar manner, the XMANet(EfficientNetB0) model exhibits enhancements of $0.21\%$ relative to the baseline EfficientNetB0, $16.48\%$ over the baseline MobileNetv3. 

Further comparisons indicate that the XMANet(MobileNetV3) model exhibits enhancements of $2.81\%$ over the baseline MobileNetv3. Additionally, the XMANet(DenseNet121) model demonstrates gains of $6.67\%$, $17.23\%$, $36.80\%$, and $4.11\%$ over the baseline ResNet50, EfficientNetB0, MobileNetV3, and DenseNet121 models, respectively.\textbf{In summary}, the proposed XMANet shows an improvement of approximately $5.04\%$ , $4.11\%$ and $2.81\%$ over the baseline ResNet50,DenseNet121, and MobileNetV3 models. These enhancements underscore the effectiveness of the proposed modifications in boosting model performance.

\begin{table}[ht]
\centering
\caption{Gesture recognition from Wavelet Transform based stacked scalogram Images using Deep learning CNN-based baseline and proposed XMANet.}
\label{tab:WTimagefors}
\resizebox{\columnwidth}{!}{%
\begin{tabular}{l|c|c|c|c}
\hline
\textbf{Models} & \textbf{ACC} & \textbf{P} & \textbf{R} & \textbf{F1} \\ \hline
ResNet50   & 54.03  &  0.54    & 0.54   &   0.54         \\
MobilNetV3   &   47.22 &   0.48   &  0.47  &  0.47         \\
DenseNet121  &       55.31    &   0.57         &   0.55         &     0.56       \\
EfficientNetB0  &       49.61     &   0.50          &   0.50         &     0.49        \\ 
XMANet(ResNet50)  &      \textbf{ 56.33 }     &  0.56          &   0.56        &     0.56       \\
XMANet(DensNet121)  &      \textbf{ 60.49 }    &   0.61         &   0.60         &     0.60      \\
XMANet(MobileNetV3)  &      \textbf{ 49.92 }     &  0.50         &   0.50        &     0.50       \\
XMANet(EfficientNetB0)  &      \textbf{ 52.63 }    &   0.53        &   0.53        &     0.53      \\
\hline
\end{tabular}%
}
\end{table}

Table~\ref{tab:WTimagefors} presents the performance metrics for various deep learning models applied to gesture recognition using wavelet transform-based stacked scalogram images FORS EMG Dataset. The experimental evaluation indicates that the baseline models ResNet50, MobileNetV3, DenseNet121, and EfficientNetB0 achieved accuracies of $54.03\%$, $47.22\%$, $55.31\%$, and $49.61\%$, respectively. With the proposed XMANet, the performance improved to $56.33\%$, $49.92\%$,$60.49\%$, and $52.63\%$ compared to baselines ResNet50, MobileNetV3, DenseNet121, and EfficientNetB0 respectively  demonstrating a clear performance enhancement.

A detailed comparative analysis shows that \textbf{XMANet(ResNet50)} exhibits a moderate improvement over its ResNet50 baseline (a $4.26\%$ increase) and the DenseNet121 baseline (a $1.84\%$ increase), with a substantial gain over the MobileNetV3 baseline (a $19.29\%$ increase) and an improvement over the EfficientNetB0 baseline (a $13.55\%$ increase). In addition, \textbf{XMANet (DenseNet121)} delivers robust performance by achieving improvements of $11.96\%$ over ResNet50, $28.12\%$ over MobileNetV3, $9.36\%$ over DenseNet121, and $21.95\%$ over EfficientNetB0. Moreover, \textbf{XMANet (MobileNetV3)} shows a slight improvement over the MobileNetV3 baseline (a $5.72\%$ increase) and nearly matches the EfficientNetB0 baseline (a $0.62\%$ increase). Meanwhile, \textbf{XMANet(EfficientNetB0)} outperforms the MobileNetV3 baseline by $11.46\%$ and records a modest gain of $6.09\%$ over the EfficientNetB0 baseline.

In summary, the proposed XMANet exhibits an improvement of approximately $4.26\%$, $9.36\%$, $5.72\%$ and $6.09\%$ over the baseline ResNet50, DenseNet121, MobileNetV3 and EfficientNetB0 model. These enhancements underscore the effectiveness of the proposed modifications in boosting model performance.

\section{Conclusion}

This study proposed a novel pipeline for raw EMG analysis that converts signals into stacked Short‑Time Fourier Transform (STFT) spectrograms and time–frequency scalograms, thereby capturing temporal dynamics with greater fidelity.  Building on these enriched representations, we introduced \textbf{XMANet}, a fine‑grained deep network that employs cross‑layer mutual‑attention learning to boost gesture‑recognition accuracy. 

Across the GrabMyo benchmark, XMANet consistently surpasses its baseline counterparts.  With STFT, the model delivers gains of roughly $1.72\%$, $4.38\%$, $5.10\%$, and $2.53\%$ over ResNet50,
EfficientNetB0, MobileNetV3, and DenseNet121, respectively.  On wavelet‑based, it records additional improvement of about $1.57\%$, $1.88\%$, $1.46\%$, and $2.05\%$ against the same baselines. For the FORS‑EMG dataset, STFT experiments show XMANet(ResNet50) improved its baseline by $\sim\!5.04\%$,  while XMANet(DenseNet121)  and XMANet(MobileNetV3) improve by approximately $4.11\%$ and $2.81\%$, over its baselines respectively.  Using wavelet inputs, the proposed architecture secures further boosts of about $4.26\%$, $9.36\%$, $5.72\%$, and $6.09\%$ over ResNet50, DenseNet121, MobileNetV3, and EfficientNetB0, respectively.

As a part of future work, a fairness‑aware version of the proposed model will be implemented to evaluate and mitigate performance bias across demographic groups. Further, advanced deep learning models will be explored to capture both local and global patterns in time–frequency analysis and leveraging self‑supervised learning for rich signal representations from unlabeled datasets.


\bibliographystyle{elsarticle-num}

\bibliography{refrences}






\end{document}